# A Target Classification Decision Aid

Todd Michael Mansell[†]
Defence Science and Technology Organisation
PO Box 4331, Melbourne, Australia.
mansell@drea.dnd.ca

## Abstract

A submarine's sonar team is responsible for detecting, localising and classifying targets using information provided by the platform's sensor suite. The information used to make these assessments is typically uncertain and/or incomplete and is likely to require a measure of confidence in its reliability. Moreover, improvements in sensor and communication technology are resulting in increased amounts of on-platform and off-platform information available for evaluation. This proliferation of imprecise information increases the risk of overwhelming the operator. To assist the task of localisation and classification a concept demonstration decision aid (Horizon), based on evidential reasoning, has been developed. Horizon is an information fusion software package for representing and fusing imprecise information about the state of the world, expressed across suitable frames of reference. The Horizon software is currently at prototype stage.

## 1. INTRODUCTION

The combat system is an integral part of the command and control of a naval vessel as it is responsible for the collection, processing and transmission of this information. Recent advances in sensor technology and communications have seen a dramatic increase in the amount of data available for processing. This translates to an increase in the amount of information available to be evaluated by command (e.g., target classification by the sonar supervisor). Often this information is uncertain, incomplete and inconclusive and is likely to include a measure of confidence in the reliability of its source (see section 2). In target classification and threat assessment one must use all available information to determine the target's identity and capabilities. To reduce the risk of information overload and assist the sonar supervisor in performing classification in an accurate and timely manner, a concept demonstration decision aid (Horizon) has been developed.

Horizon is based on a methodology for representing and reasoning with information from disparate sources, expressed across a number of frames of reference, called evidential reasoning (Lowrance *et al* 1991). Horizon provides an environment for the operator to propagates and fuse the initial information to produce a measure of confidence in a target's classification. The initial information can be translated, discounted or fused, using one of three algorithms, in a graphical manner. Horizon is a concept demonstration software package that is also being applied to mine threat evaluation (Mansell 1996) and air combat post mission analysis (Mansell 1997) domains.

This paper discusses the information fusion problem as it applies to target classification and the development of an information fusion decision aides (Horizon). Horizon draws on the established evidential reasoning techniques and adds a new fusion algorithm for combining dependent evidence, as well as automated discounting and explanation facilities. The paper also highlights some of the practical issues associated with designing an information fusion system to be used by non-technical domain experts (Navy personnel).

## 2. TARGET CLASSIFICATION

The process of target classification has numerous definitions in navies around the world. These definitions essentially provide a methodology for deriving a contact description at one of the following classification levels; (1) general classification (e.g., submarine) (2) vessel type (e.g., SSK) and operation condition, (3) nationality and class (e.g., Canadian Oberon class), or (4) particular unit (e.g., Onondaga). The target classification problem should not be considered a finite process; rather it is the derivation of essential contact information from a dynamic environment. The operator's overall task in this process is to participate with command in target detection, localisation, and identification (i.e., target acquisition). This is achieved by the operator interpreting the data presented on his screen(s), and translating that into informative reports to command to permit tactical decisions to be made (Donald, 1996a, b).

[†] Currently on exchange at Defence Research Establishment Atlantic PO Box 1012, Dartmouth, NS, B2Y 3Z7, Canada.

To assist the operator in target classification a decision aid must be sufficiently flexible to reason at all levels of classification (be it general classification, type, nationality, class, or unit) using the information as it becomes available. The operator will use the decision aid to continuously evaluate the available evidence and report classification information to command. The level of classification in the report depends on the quality of information and current operational conditions.

To further complicate the classification process, the operator must deal with information that is derived from an environmentally hostile medium (ie the ocean). As an example, sound transmission often occurs in surges and fades, and rarely will all radiated frequencies be detectable at the same time and place. It is the sonar supervisor's task to consider all the available, imprecise sensor and intelligence information and to provide a classification for sonar contacts. That includes information from the following sources:

- Raw bearing only information interpreted by the acoustic (sonar) operators.
- Individual tone information provided by processing the radiated sounds in narrow frequency bands (e.g., gear tonals).
- Harmonic sets information as it relates to machinery on the target platform.
- Transients information interpreted by the acoustic sensor operator or from automated transients analysis system.
- Detection of emissions from the target vessel's active sonar.
- Aural classification using the experience of the sensor operator.
- Visual information provided by the periscope.
- Electronic support measures (ESM) information (e.g., radar emissions).
- Radar.
- Intelligence identifying what one expects to find in the area.

Horizon has been developed as a flexible decision aid that can reason with imprecise information from the above sources. Horizon uses a unique combination of existing and novel evidential reasoning techniques to provide a mathematically rigorous formalism for combining bodies of evidence at different levels of abstraction (see the following section).

## 3. EVIDENTIAL REASONING IN HORIZON

Evidential reasoning (Lowrance *et al* 1991) is a formalism for representing information from disparate sources that is expressed in different frames of reference, and provides techniques to manipulate that information. Evidential reasoning (E-R) is an extension of Shafer's (1976) work on belief functions (called Dempster-Shafer Theory). Being a departure from classical probability theory, E-R uses information that is typically uncertain, incomplete and error-prone. E-R maintains the association between the measure of belief and disjunctions of events rather than forcing probabilities to be distributed across atomic possibilities.

E-R is used to assess the effect of all pieces of available evidence on a hypothesis, making use of domain-specific knowledge. A propositional space called the *frame of discernment* (or frame) is used to define a set of basic statements, exactly one of which may be true at any one time, and a subset of these statements is defined as a *propositional statement*. For example, a frame, $\theta_A$, may be used to represent a targets category (this implementation of the target classification domain currently uses 16 frames to describe the environment as shown in *Figure 1*).

Once frames have been established, basic probability assignments (BPA) are use to make probabilistic assessments about the confidence in propositional statements relative to the frame. Belief assigned to non-atomic propositional statements explicitly represents the lack of information available to resolve between the propositions, resulting in a distribution appropriate to the granularity of the evidence. The term body of evidence (BOE) is used to describe the unit distribution of BPAs over propositional statements discerned by a frame of reference and in accordance with the information source.

E-R provides a complete methodology for information integration, including the collection of information in its native frame of reference, discounting (due to the credibility of the source), translation to a related frame, projection into the future (or past), and fusion with other independent BOEs.

Compatibility relations are used to characterise interrelationships between different propositional spaces. This allows reasoning to be carried out on information described at different levels of abstraction or on frames of reference with overlapping attributes. *Figure 1* shows all the frames used in the target classification domain, with a link between two frames representing the existence of a compatibility relation. In this domain, a compatibility relation between the *classification* and *diesels* frames, represents the number of diesel motors known to be available on each platform class. Therefore, evidence about the number of diesel motors observed provides information about the type of platform.

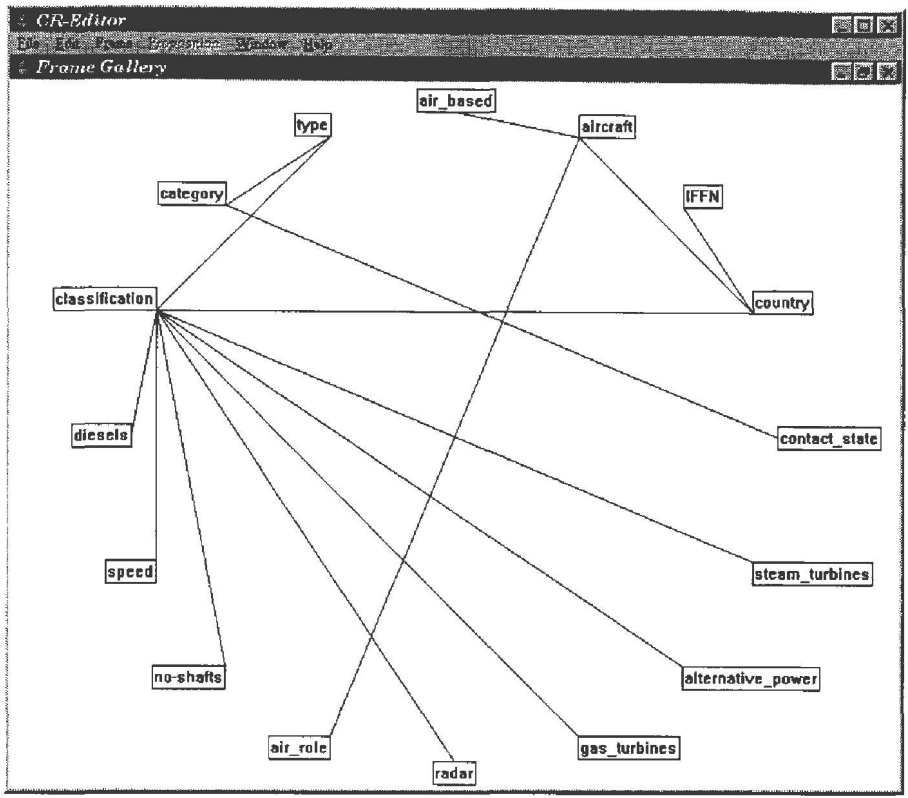

Figure 1: Sample CR-Editor window showing the frames that are linked by compatibility relations in the classification domain.

E-R uses Dempster's rule of combination (Lowrance *et al*, 1991) to fuse multiple independent BOEs into a single BOE, emphasising points of agreement and deemphasising points of disagreement. Dempster's rule is both commutative and associative (ie, evidence can be combined in any order) providing a consensus of what was disparate opinion. Alternative fusion algorithms have also been proposed to counter perceived weaknesses of Dempster's rule. Horizon has implemented Dempster's rule as well as Smets algorithm (Smets, 1993) and we are currently evaluating their strengths and weaknesses. Initial results suggest Smets algorithm may suit this military problem as it provides a conservative weighting of evidence. Further, Smets method of distributing conflicting evidence to the *unknown* proposition (which states the true proposition may not be an element of the frame) suits the dynamic military domain where weapons and platforms are rapidly evolving. Users may then be trained to recognise that a high value for the unknown proposition may suggest (1) a high level of conflict in the evidence (2) signature data for a target is not included in the database, or (3) the target may be deliberately masking its identity by altering its recognisable signatures.

The selection of this method for dealing with uncertainty was not based on competency, as probability theory and fuzzy logic are very capable of representing uncertain information. Instead, E-R was eventually selected over probability theory for its natural representation and manipulation with information contained at different levels of abstraction (Gordon and Shortliffe, 1985 and Almond, 1995) and in different frames of reference.

### 3.1 INDEPENDENCE OF EVIDENCE

The concept of independence is controversial in the areas of E-R and probability theory (Dawid 1979). Often centring around the areas of experimental independence or conditional independence, these theories have tended to handle dependence inadequately (Pearl 1988, Shafer 1981, Walley 1991, Kahneman, *et al* 1982). These criticisms are usually based on the difficulty of acquiring the appropriate evidence values, and applying an independence test to the data.

It has been proposed that conditional independence between BOEs is not sufficient to guarantee the validity of Dempster's fusion algorithm (Voorbraak 1991). However, these papers fail to provide a decisive argument, or unbiased counter-intuitive example, that proves conditional independence combined with screening of BOEs by a domain expert is insufficient for guaranteeing independence of evidence. Hence, when using Horizon one makes the following two assumptions about the BOEs being fused using Dempster's rule:

- the human operator can determine whether BOEs are based on the same observations, and are therefore dependent (e.g., two intelligence reports quoting the same source are not independent).
- people (i.e., the expert and knowledge engineer) can accurately and confidently determine whether two events or actions are independent, as proposed by Pearl (1988), Shafer (1976, and 1981), Walley (1991), and Spiegelhalter and Lauritzen (1990).

There is a growing number of successful applications of E-R to real world problems in the literature, including submarine tracking, sonar data interpretation, anti-air threat identification and naval intelligence analysis (Lowrance *et al* 1991). The existence of these positive experiences demonstrate a level of maturity of the technology and implies, at least empirically, that dependent BOEs are being identified (or perhaps weak dependencies that are not identified have minimal effect on the conclusions).

## 4. THE HORIZON PROGRAM

The software package called Horizon1 (currently at the prototype stage) is a domain-independent E-R system that has been applied to the mine threat evaluation domain and is currently being applied to the target classification problem. One challenge in developing this information-fusion software package is to make sure the design does not require the user to understand the intricacies of E-R (a goal we are still working towards). However, it is anticipated that a certain amount of understanding (training) is required to distribute evidence in an E-R manner, as well as ensure the information is independent.

Horizon is a decision-aid program that requires a knowledge engineering process to take place before it can be applied to a problem. This involves capturing the domain by first establishing the frames of reference used to represented BOEs, and generating the compatibility relations between those frames. The amount of knowledge engineering required will depend on the domain under investigation. The target classification domain consists of 16 frames of reference used to describe different attributes of a target. The compatibility relations are constructed with the aid of an expert, and represent the frames of reference in which one expects information to arrive, or may require conclusions to be presented.

Knowledge engineering of the target classification domain presented in this paper was a straightforward, albeit laborious, task. The relationships between platform classification and its various components (e.g., number of shafts) were extracted from reference material such as *Jane's Fighting Ships* (Sharpe, 1995). This domain describes the ships and submarines from a selection of countries in the pacific theatre; which includes large and small navies such as the USN and Papua New Guinea's Navy respectively.

Horizon provides a graphical user interface for the real time display and editing of compatibility relations, called the CR-Editor. The CR-Editor has two types of windows. The *Frame Gallery* window links frames for which compatibility relations exist (displayed in *Figure 1*). The *CR-windows* display the compatibility relations between two frames of reference with the propositions of each frame lined up on each side. Links between these propositions represent information in one frame of reference that is simultaneously true in the other frame of reference. For example, the Oberon class submarine (from the classification frame) in Horizon is linked to:

- Australia and Canada (country frame);
- SSK (type frame);
- 17 (Speed frame in kts);
- 2 (number of diesels frame);
- 1 (number of shafts frame).

The CR-Editor was used to aid in the knowledge engineering of the target classification domain, and would be used to update the compatibility relations as navies acquire new assets or upgrade current assets.

Information collected from the environment (section 2) can be entered into the system in three ways. Firstly, static information (such as a database of platforms) that does not change rapidly can be stored as BOE data files, and read into Horizon when the system is initialised. Secondly, dynamic information can be entered automatically into Horizon's database from the combat system (e.g., sensors, signal processing units, expert systems, etc.). Finally, other dynamic information (such as surveillance reports) can be entered directly into the system as it arrives using the user interface window shown in Figure 2. This requires the user to select the frame of reference, then distribute belief among the listed propositions. The interface window is also used to edit and update all forms of information when required.

---

[1] Horizon is defined in Webster's Dictionary as the fullest range or widest limit of perception, interest, appreciation, knowledge, or experience.

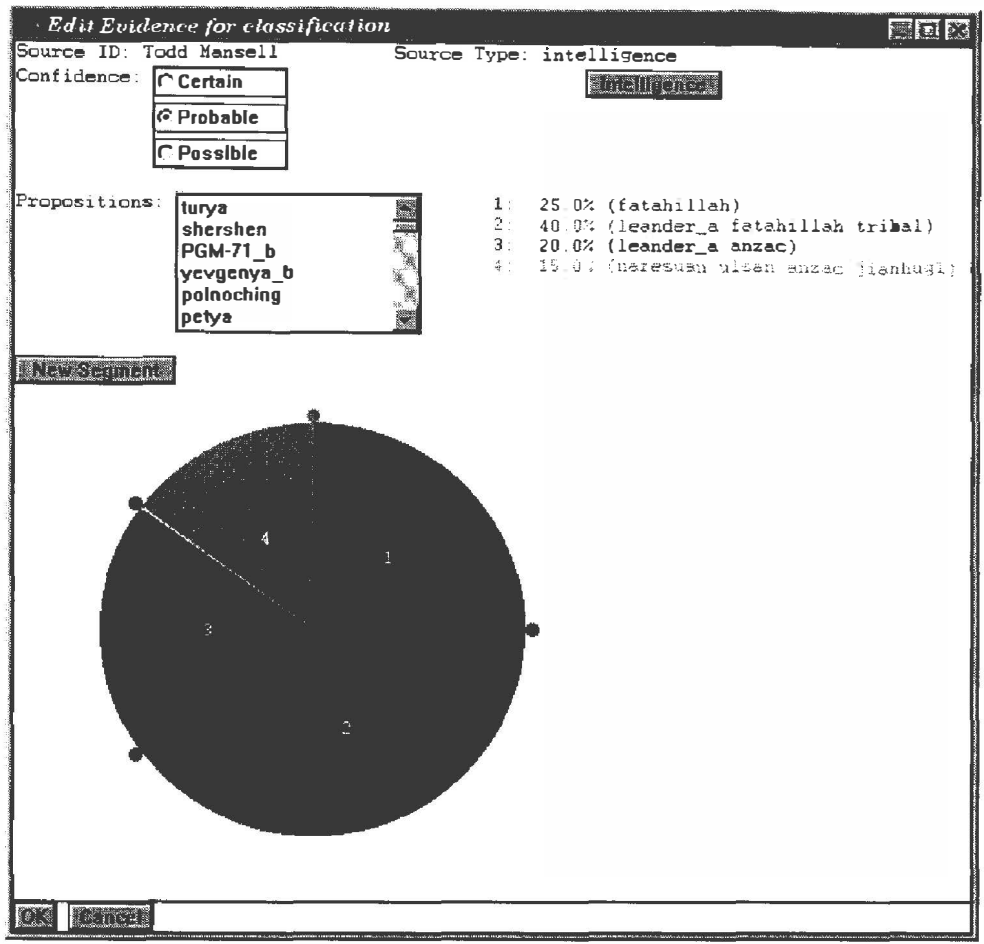

Figure 2: Windows used to create or edit a new BOE.

Horizon represents and manipulates BOEs in an object oriented manner (see Figure 3). Each BOE represented by an icon and is stored in its native frame of reference, where it can be selected to be included in a calculation. At present the calculation operations include *discount* (reduces the confidence in a BOE), *translate* (move to a new frame), and *fuse* (combine BOEs using one of three fusion algorithms). Once the user has selected the BOEs to be included in the calculation, the operation is chosen. If *discount* is selected the user supplies a discount rate (a percentage between 1 and 100) at which time Horizon produces a secondary[2] BOE with a modified belief distribution. If the *translate* operations are selected, the user is prompted to choose the frame in which the conclusion should be expressed and Horizon generates the new BOE by performing the minimum number of intermediate translations.

The operator may also chooses to perform a fusion operation using either Dempster's rule of combination (Shafer 1976), a least commitment algorithm (Mansell 1997), or a dependent evidence algorithm (Mansell 1997). The system will carry out the necessary translations, presenting the resulting BOE in the conclusions window (Figure 4). The display window presents the pooled evidence for and against all non-zero propositional statements, as well as a measure of uncertainty (being the amount of evidence that neither supports nor contradicts that statement).

Horizon is written in Allegro Common Lisp, with the user interface being written in PC/CLIM, making the package portable to either PC or UNIX machines. Horizon's "point-and-click" interface has been commended for its easy of use and provision for rapid interpretation of information (a critical feature of Naval systems).

### 4.1 AUTO-DISCOUNTING

Horizon also introduces the concept of auto-discounting as a way to condition each BOE prior to fusion. This uses the BOE's measure of confidence in the source and its ability to accurately interpret the environment. Auto-discounting (as the name suggests) is the automatic discounting of intelligence information and sensor reports based on their credibility. In practice, this requires that each BOE is assigned a measure of confidence in the source, and is initiated only after the user begins a fusion operation.

---

[2] A secondary BOE is a body of evidence that is generated through the manipulation of initial BOE(s).

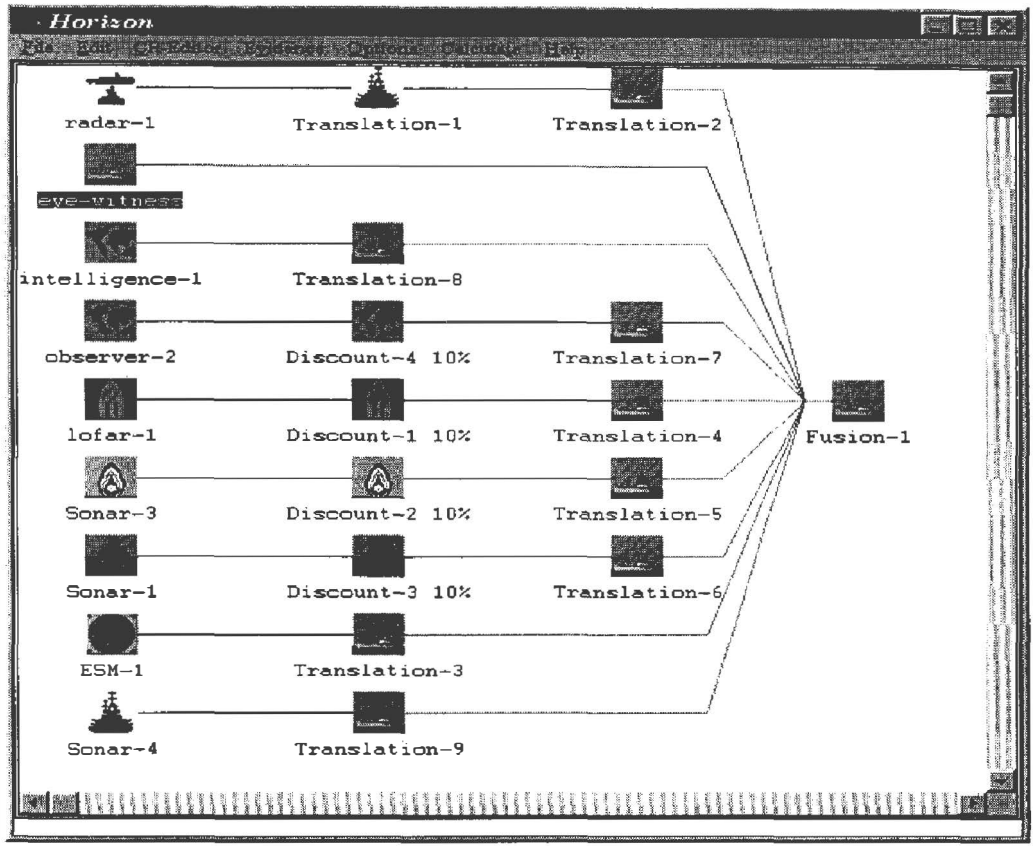

Figure 3: Window displaying the fusion of multiple BOE.

To be consistent with Australian Navy terminology, Horizon uses three levels of confidence in an information source; *certain, probable* and *possible*. Horizon requires each BOE be given a measure of confidence in its source, the default being *probable*. Each of these measures represents a discount rate for a discount operation. These rates are user configurable[3] and are currently set at:

- Certain: discount rate = 0% (representing complete confidence in the source).
- Probable: discount rate = 20% (moderate confidence in the source).
- Possible: discount rate = 40% (low confidence in the source).

The function of auto-discounting is an option that can be selected or de-selected prior to the execution of any operation, and takes place only once at the beginning of any fusion operation (ie, prior to translation). The confidence level is set from the *Edit-BOE* window using a check box with the three options *certain, probable,* and *possible* (see Figure 2).

The inclusion of an auto-discounting feature in Horizon serves two purposes. First, automatic discounting of evidence by the identity of its source allows us to minimise personal biases[4] contained in BOEs provided by humans. In addition, a sensor's supervisor unit or horizon's user may provide this measure of confidence in a BOE based on recent history or the conditions under which the information was acquired. A Second reason for introducing the measure of confidence in a source's credibility is to circumvent the Zadeh objection (Zadeh, 1984). Zadeh (1984) showed that if a propositional statement has zero or very low BPA in one BOE, then that propositional statement will always have zero or low support when fused with other BOEs, irrespective of the level of support. This occurs because when multiplying one number by a zero or very low number, one always produce a zero or very low number. By using these additional confidence measures the auto-discounting routine redistributes the BPAs resulting in a moderate (not zero or very low) BPA for the set of all propositions. Hence, if one BOE incorrectly overlooks the possibility of a proposition being true, it may still gain significant support from other BOEs after fusion.

---

[3] Investigations into the appropriateness of these values is ongoing, but initial results are positive. Most of the debate here has been whether Certain should have a zero discount rate, or a very low value (e.g., 5%).

[4] In my experience, uniformed personnel as a whole tend to overestimate quantitative measures of belief, while conservatively estimating their confidence in that measure.

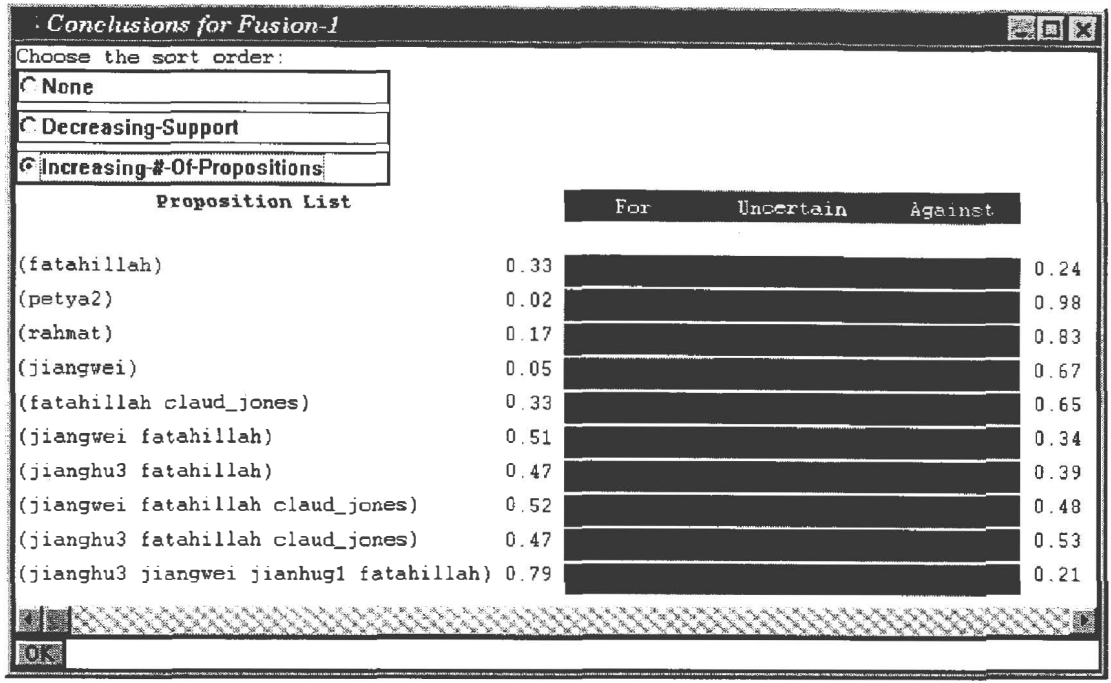

Figure 4: Window used to display the result of fused BOEs. This window presents the amount of evidence supporting, uncommitted (uncertain), and contradicting all non-zero propositional statements.

One drawback to auto-discounting is the information content of the BOE is lowered proportional to a the measure of confidence (ie, the associated discount rate). The consequence of this redistribution is that results may become inconclusive. This is particularly true when initial information quality is bordering on the inconclusive (i.e., one proposition, or set of propositions, is only slightly favoured over the others after a standard Dempster's Rule fusion). Under these circumstances auto-discounting is not recommended. To counter this, the operator could be trained to use auto-discounting by default. However, when inconclusive result are produced he/she could be trained to redo the fusion without auto-discounting.

### 4.2 SENSITIVITY ANALYSIS

An essential part of decision aid software is providing the user with some way to review or understand how the conclusion was derived (ie, an explanation facility). Endowing a system with such a feature contributes to the users confidence in the system by making it more transparent. This also provides him/her the opportunity to analyse the decision process and perhaps restructure and re-evaluate the problem. Horizon generates an explanation based on the *measure of information,* I(bel), reported in Xu and Smets (1996):

$$I(bel) = -\sum_{a \subseteq \theta} \log q(a),$$

where q(a) is the commonality function:

$$q(a) = \sum_{a \subseteq B \subseteq \theta} m(B).$$

Horizon's explanation algorithm also relies on the additive property of this measure of information. That is, $I(bel_{12})=I(bel_1)+I(bel_2)$, where $bel_{12}$ is the belief function that results from the combination of $bel_1$ and $bel_2$ using Dempster's unnormalised rule. The full implementation of this algorithm and has been reported in Xu and Smets, (1996) and has been adapted to apply to Mansell's (1997) dependent evidence fusion rule. Those BOEs most influential on the conclusion are identified by highlighting the links in the main window (Figure 3). A qualitative explanation may also be displayed in the form of a text window that identifies the most and least influential BOEs.

In Figure 3 the *Eye-Witness* BOE had the most impact on the conclusion (signified by a red link to the conclusion BOE). The user may choose to discount or remove *Eye-Witness* and recalculate the conclusion if his/her confidence in the BOE is in question.

## 5. DISCUSSION

Initial results from the target classification domain indicate that independence of evidence is not a serious concern for most sources. However, this may not be the case when dealing with intelligence reports, as it is often difficult to determine independence of sources (particularly when the basis of the reports are unknown). Horizon includes an algorithm to fuse suspected dependent BOEs and the utility of this function is under investigation.

Horizon has been trialed on a limited set of synthetic data to demonstrate the suitability of evidential reasoning to information fusion. On this data (up to 20 BOEs) the fusion runs at near real time, with frames containing between 3 and 15 propositions, (and an average size of 8.6). Preliminary results for a worst case computation show that for 35 BOEs with between 8 and 352 (average 211) propositions, Horizon takes 2.3 minutes to perform 25 discounting operations, 29 translation operations and 35 fusion operations on a Pentium 133. This time frame is considered reasonable for some tactical situations (e.g., target classification onboard a submarine). However, Horizon is a technology demonstrator and is therefore not optimised to be fast or efficient.

## 6. CONCLUSION

This paper describes the target classification problem and the Horizon decision aid currently under development. Preliminary examination of this software confirms that evidential reasoning is an appropriate technique for dealing with high level information fusion. Further, the methodology is sufficiently mature to allow the development of robust decision systems that can accurately fuse and display tactical information.